\documentclass[journal]{IEEEtran}

\usepackage[utf8]{inputenc}
\usepackage[T1]{fontenc}
\usepackage{microtype}

\usepackage{amsmath}
\usepackage{amssymb}
\usepackage{amsfonts}
\usepackage{mathtools}
\usepackage{bm}
\usepackage{amsthm}

\usepackage{graphicx}
\usepackage{booktabs}
\usepackage{multirow}
\usepackage{array}

\usepackage{xcolor}
\usepackage{enumitem}

\usepackage{algorithm}
\usepackage{algpseudocode}

\usepackage{hyperref}
\usepackage[nameinlink]{cleveref}

\graphicspath{{figures/}{images/}}

\hypersetup{
    colorlinks=true,
    linkcolor=blue,
    citecolor=blue,
    urlcolor=blue
}

\theoremstyle{definition}

\theoremstyle{remark}

\newcommand{\R}{\mathbb{R}}

\newcommand{\HH}{\mathcal{H}}
\newcommand{\Hk}{\mathcal{H}_k}
\newcommand{\F}{\mathcal{F}}
\newcommand{\KL}{\mathrm{KL}}

\newcommand{\norm}[1]{\left\lVert #1\right\rVert}
\newcommand{\inner}[2]{\left\langle #1,\, #2\right\rangle}

\title{Globalized Constrained Stein Variational Inference for Diverse Feasible Robot Motion Planning}

\author{
Jiayun~Li$^{1,2}$, \, Georgia~Chalvatzaki$^{1,2,3}$%
\thanks{$^{1}$PEARL Lab, Dept. of Computer Science, TU Darmstadt.}%
\thanks{$^{2}$Hessian.AI, Darmstadt, Germany.}%
\thanks{$^{3}$Robotics Institute Germany (RIG).}%
}

\begin{document}

\maketitle

\begin{abstract}
Robot motion planning is inherently multimodal, yet classical planners typically return only a single solution. Probabilistic formulations address this limitation by maintaining a distribution over motions, allowing the planner to reason over multiple low-cost alternatives. In robotics, however, motion samples must also satisfy strict constraints, including collision avoidance, joint limits, contact conditions, and dynamics consistency. These hard requirements make motion sampling substantially more challenging: within a limited planning budget, the ensemble must cover diverse low-cost motions while ensuring that every sample remains feasible under the relevant constraints.
We propose SteinSQP (\textit{Stein Variational Sequential Quadratic Programming}), a constrained Stein variational inference method for diverse feasible robot motion sampling. SteinSQP evolves an interacting particle ensemble, as in Stein variational methods, while embedding constraints directly into a kernel-space SQP subproblem. We solve the resulting constrained Stein--Newton subproblem with a GPU-friendly matrix-free primal--dual algorithm, enabling efficient batched ensemble updates. To globalize the method, we introduce an ensemble-level merit function that jointly balances objective value, constraint violation, and particle diversity.
Across five constrained motion-planning tasks, SteinSQP returns fully feasible ensembles while preserving diverse motion alternatives. Compared with first-order constrained Stein baselines and serial multistart nonlinear programming, SteinSQP shows faster and more robust ensemble convergence in terms of iterations, improves particle-wise feasibility, and achieves faster batched time-to-solution on challenging robot-scale tasks.
\end{abstract}


\section{Introduction}
\label{sec:introduction}

Robot motion planning is conventionally formulated as the search for a single trajectory that minimizes a cost functional subject to hard constraints such as joint limits, collision avoidance, and contact conditions~\cite{ratliff2009chomp, schulman2014motion}. Yet a single minimizer discards structure that most tasks naturally possess: a manipulator can pass an obstacle on either side, a throw can be released from different arm configurations, and a push can exploit different contact faces. A planner committed to one solution is brittle precisely where planning matters most. It is easily trapped in a poor local optimum, and it offers no fallback when the environment changes, when a downstream module rejects the trajectory, or when execution fails. What is needed instead is a \emph{set} of low-cost, feasible, and qualitatively distinct motions, produced within a single planning budget.

Casting motion optimization as probabilistic inference provides precisely this structure~\cite{toussaint2009robot, mukadam2018continuous}. The objective induces a distribution over the optimization landscape, with high-density regions concentrated near distinct local optima of the underlying nonlinear program, while hard constraints restrict its support to the feasible set: a manifold under equality constraints, or a domain with boundary under inequalities. Sampling from this constrained distribution therefore returns a diverse collection of feasible motions in one solve, rather than a single point estimate. The difficulty is computational. The samples must be simultaneously low-cost, feasible with respect to nonlinear equality \emph{and} inequality constraints, diverse, and they must be obtained fast enough to be useful for planning.

\begin{figure}[t]
    \centering
    \includegraphics[width=0.95\linewidth]{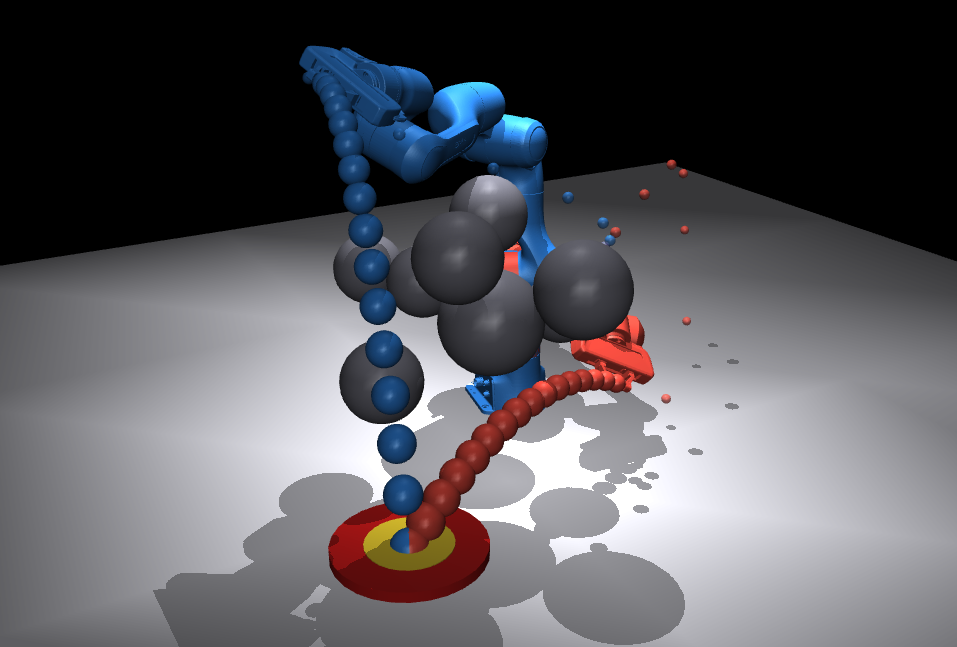}
    \caption{SteinSQP samples a diverse set of feasible throwing motions in a single solve: the arm and ball avoid the obstacles (black) and release the object along a ballistic arc onto the target, while the surrounding particles trace the in-hand trajectories of the ball before release.}
    \label{fig:panda_throw}
\end{figure}

Classical constrained samplers fall short of this requirement because they are inherently sequential. Manifold MCMC methods~\cite{brubaker2012family} are asymptotically exact, but rely on long mixing chains and do not share information across samples. Constraint-manifold RRTs~\cite{berenson2009manipulation, suh2011tangent} similarly expand one projected node at a time. Neither maps well onto the parallel hardware that modern robot software stacks are built around. Stein Variational Gradient Descent (SVGD)~\cite{liu2016stein} does. It evolves an entire ensemble of interacting particles in parallel, each driven toward low cost while a pairwise repulsion prevents collapse onto a single solution: in effect, a batched multi-start optimizer whose restarts communicate. SVGD, however, assumes an unconstrained domain. Its constrained extensions either apply only to restricted constraint classes, namely convex sets via mirror maps~\cite{shi2021sampling} or a single equality manifold via projected kernels~\cite{zhang2022sampling}, or handle nonlinear inequalities indirectly, through squared slack variables~\cite{power2024constrained} or penalty and augmented-Lagrangian terms~\cite{tabor2025constrained}. These reformulations share a common defect: they alter the particle flow precisely at active constraint boundaries, where feasibility is most delicate, introducing spurious local optima and distorting the empirical distribution of the ensemble exactly where the constrained posterior concentrates.

In this article we take a different route: rather than reformulating the constraints, we make each Stein step feasible by construction. Inspired by the interpretation of constrained SVGD as SQP~\cite{li2025constrained} and by safe particle flows~\cite{yi2025constrained}, we linearize the constraints at the current ensemble and enforce them directly on the kernel-coupled ensemble displacement via a quadratic program (QP). This lets equality and inequality constraints enter the particle flow at first order, without slack variables or hand-tuned penalties. Three observations turn this idea into a practical solver. First, the same QP can incorporate second-order curvature information, such as Gauss--Newton approximations, upgrading the safe first-order flow to a constrained Stein--Newton update. This improves convergence speed while preserving constraint satisfaction, rather than merely enforcing feasibility. Second, although the kernel coupling makes the QP's KKT system dense and prohibitive to factorize, a primal--dual hybrid gradient scheme (PDHG)~\cite{chambolle2011first, goldstein2013adaptive} that never forms this system solves it using only kernel products and small independent per-particle solves, a computational pattern that GPUs execute almost natively. Third, a QP step trusts a linearization and must therefore be globalized. Prior constrained Stein methods either omit globalization entirely or line-search on objective alone, which collapses the very modes the ensemble exists to preserve. We instead globalize with an \emph{ensemble-level} merit function that balances the three goals a constrained sampler must serve simultaneously: low cost, constraint satisfaction, and transport diversity. By tailoring globalization to the Stein particle flow, this mechanism safeguards ensemble diversity while steering particles toward feasible, low-cost regions.

We call the resulting method \textbf{SteinSQP}: an SQP-like particle optimizer that turns constrained Stein transport into a practical GPU-parallel solver for multimodal nonlinear programs. We summarize our contributions as follows:
\begin{itemize}
\item \textbf{Constrained Stein--Newton updates in kernel space.}
We formulate an SQP-like Stein variational Newton step that embeds linearized equality and inequality constraints directly into the kernel-coupled ensemble displacement, avoiding slack variables, fixed penalties, and post-hoc projection.
\item \textbf{A GPU-parallel primal--dual solver.}
We solve the dense kernel-coupled QP without forming or factorizing its KKT system, using only kernel products and independent per-particle solves. The resulting dual iterates also provide the exact-penalty weight used by the merit function, eliminating a fragile hand-tuned parameter.

\item \textbf{Stein-aware globalization.}
We introduce an ensemble-level merit function for constrained Stein updates that jointly accounts for cost, feasibility, and transport diversity, together with a second-order correction to mitigate Maratos-type stalling.

\item \textbf{Empirical validation on robot-scale tasks.}
Across five benchmarks, SteinSQP achieves fully feasible ensembles while improving robustness and batched wall-clock time over both first-order Stein and serial multistart NLP baselines. It also shows a faster outer-iteration convergence than serial multistart NLP.
\end{itemize}

The remainder of this article is organized as follows. Section~\ref{sec:related} reviews constrained sampling for motion planning and constrained Stein variational inference. Section~\ref{sec:problem_definition} formalizes the constrained posterior sampling problem, and Section~\ref{sec:prelim} recalls the necessary background on kernel transport maps and Stein variational inference. Section~\ref{sec:method} develops SteinSQP, Section~\ref{sec:experiments} presents the experimental evaluation, and Section~\ref{sec:conclusion} concludes.
\section{Related Work}
\label{sec:related}

\subsection{Constrained Sampling for Robot Motion Planning}
Casting motion optimization as probabilistic inference lets a planner reason
about a \emph{distribution} of low-cost motions rather than a single
minimizer~\cite{toussaint2009robot,mukadam2018continuous}. When hard constraints
are present, this posterior is supported on a lower-dimensional feasible set,
and the central difficulty becomes drawing \emph{diverse} samples
that remain feasible. Sampling-based planners address this geometrically:
projection-based methods repeatedly retract configurations onto the constraint
manifold~\cite{berenson2009manipulation}, while Tangent Space
RRT~\cite{suh2011tangent} and Tangent Bundle RRT~\cite{kim2016tangent}
grow trees in local tangent charts. Both incur high projection cost and
struggle with strongly nonlinear constraints. A complementary line treats the
feasible set as an implicitly defined manifold and applies \emph{manifold
MCMC}: constrained Hamiltonian Monte Carlo with RATTLE-type symplectic
integrators~\cite{brubaker2012family,lelievre2018hybrid}, Riemann
manifold HMC~\cite{girolami2011riemann}, and geodesic Monte
Carlo~\cite{byrne2013geodesic}. These schemes are asymptotically exact but rely
on sufficient chain mixing, share no information across samples, and are hard
to parallelize, which limits their use under tight planning budgets. Closer to
our setting, combining Gauss--Newton constraint satisfaction with interior
sampling produces diverse constrained samples~\cite{toussaint2024nlp}, but still
proceeds sequentially, while learned generative models amortize manifold
sampling~\cite{ortiz2022structured} at the cost of training data and poor
generalization to unseen constraints. In contrast, particle-based variational
inference evolves an interacting ensemble jointly, coupling samples through a
repulsion term so that diversity and feasibility are pursued in parallel, the
regime our method targets.

\subsection{Constrained Stein Variational Inference}
Stein Variational Gradient Descent (SVGD)~\cite{liu2016stein,liu2026probabilistic} approximates a
target by a set of interacting particles that follow the steepest
KL-descent field in an RKHS, balancing a score-driving term against a
repulsion term, and can be viewed as a gradient flow of the KL
divergence~\cite{liu2017stein}. Second-order variants precondition this
flow: the Stein Variational Newton (SVN) method~\cite{detommaso2018stein}
approximates a Newton iteration in function space, and related second-order
Stein schemes have since been applied to dynamic optimization and
control~\cite{aoyama2025second}. Extending SVGD to constrained supports has
followed three broad strategies. Mirror-map methods evolve particles in a
dual space to remain within convex constraints~\cite{shi2021sampling};
moment-constraint approaches enforce expectation constraints through a
primal--dual flow~\cite{liu2021sampling}; and projection methods build the
constraint into the function space itself. In particular, Orthogonal-space
SVGD (O-SVGD)~\cite{zhang2022sampling} uses a projected matrix-valued kernel
whose fields are tangent to a single equality manifold, while functional
gradient flows with boundary conditions confine particles to general domains
with convergence guarantees~\cite{zhang2024functional}. Further extensions, such as Riemannian SVGD~\cite{liu2018riemannian} and Lie-group SVGD~\cite{li2026stein}, are tailored primarily to intrinsic manifold constraints, but may be less straightforward to deploy in general-purpose applications.

In robotics, these ideas underpin recent constrained trajectory samplers.
CSVTO~\cite{power2024constrained} extends O-SVGD to multiple equality and inequality
constraints with a tangent-space resampling step, but handles inequalities
through squared slack variables, which, as that work notes, introduce many
spurious local optima and distort the empirical distribution near active
boundaries. Safe particle flow~\cite{li2025constrained,yi2025constrained} instead enforces linearized
inequalities via a control-barrier quadratic program, and augmented-Lagrangian
formulations of constrained SVGD handle arbitrary constraints at the cost of
penalty tuning~\cite{tabor2025constrained}. What remains open is a method that
treats nonlinear inequalities directly, without slack variables or fixed
penalties, while retaining a globalized, curvature-aware Stein flow that
parallelizes across particles. SteinSQP targets exactly this gap: each Stein
proposal is embedded in an SQP subproblem solved in kernel space and
globalized by an exact-penalty line search, with the inner saddle-point
problem handled by a matrix-free primal--dual scheme that maps well onto GPU
hardware.
\section{Problem Definition}
\label{sec:problem_definition}
A Boltzmann distribution formulation casts optimization-based robot motion optimization as probabilistic inference. Let $x$ denote a finite-dimensional motion variable, which may represent robot configuration parameters or a trajectory in generalized coordinates. Given a prior distribution $p_0(x)$, the likelihood takes the form $l(x) \propto \exp(-V(x)/T)$, where $V(x)$ denotes the motion-level cost and $T>0$ controls the temperature. This formulation yields the posterior distribution $p(x) \propto p_0(x)\exp(-V(x)/T)$. As $T \to 0$, the posterior concentrates on the minimum-cost motions within the support of $p_0(x)$, recovering deterministic motion optimization in the zero-temperature limit.

We consider a general constrained nonlinear program with a nonlinear least-squares objective $V(x)=\tfrac{1}{2}\|r(x)\|^2$, where $r:\mathbb{R}^d\to\mathbb{R}^{m_r}$ is a smooth residual map. The feasible set is defined as $\mathcal{F}\doteq \{x \mid h(x)=0,\ g(x)\leq 0\}$, where $h:\mathbb{R}^d\to\mathbb{R}^{m_h}$ and $g:\mathbb{R}^d\to\mathbb{R}^{m_g}$ denote equality and inequality constraint functions, respectively, with the inequality applied elementwise. This formulation covers a wide range of robotics motion planning tasks, such as motion planning with obstacle avoidance. Moreover, even when contact dynamics are involved, they can be handled via relaxation and smoothing techniques to enable a nonlinear programming formulation, further broadening the applicability of this formulation.

To incorporate feasibility, we introduce the extended-value indicator $\iota_{\mathcal{F}}(x)$, where $\iota_{\mathcal{F}}(x)=0$ if $x\in\mathcal{F}$ and $\iota_{\mathcal{F}}(x)=+\infty$ otherwise. The constrained Boltzmann likelihood induced by $V(x)+\iota_{\mathcal{F}}(x)$ is equivalent to $\exp(-V(x)/T)\,\mathbf{1}_{\mathcal{F}}(x)$, where $\mathbf{1}_{\mathcal{F}}(x)$ is the binary feasibility indicator. The target posterior is therefore given by
\begin{equation}
\label{eq:constrained_posterior}
\pi(x) \propto p_0(x)\exp \bigl(-V(x)/T\bigr) \, \mathbf{1}_{\mathcal{F}}(x),
\end{equation}
which assigns zero density to infeasible motions.

Our goal is to draw samples from this constrained posterior, where each sample represents a feasible robot motion and different modes correspond to distinct local optima. This perspective positions sampling not only as a means of approximating the posterior, but also as a mechanism for uncovering multiple high-quality local modes. We therefore seek a method that achieves a favorable trade-off among sampling efficiency, sample diversity, and fidelity to the constrained target distribution.

In the context of Stein variational inference, this objective requires a Stein operator that respects the constrained motion domain. We introduce the necessary preliminaries in the next section.

\section{Preliminaries}
\label{sec:prelim}

This section introduces notation and recalls the three ingredients our method
builds on: reproducing-kernel function spaces, Stein variational inference
viewed as optimization over a particle ensemble, and the enforcement of
equality constraints as a tangency condition on the ensemble's transport
field.

\subsection{Kernel Spaces and Transport Maps}
\label{sec:prelim_rkhs}

Let $k:\R^d\times\R^d\to\R$ be a positive-definite kernel and $\Hk$ its
scalar reproducing-kernel Hilbert space (RKHS);
$\HH \doteq \Hk^d$ denotes the corresponding space of vector fields
$\delta:\R^d\to\R^d$ with componentwise inner product. Two properties matter
here. First, the \emph{reproducing property}: for any $\delta\in\HH$,
$\inner{\delta}{k(x,\cdot)v}_{\HH} = \delta(x)^\top v$, so function evaluations are
inner products and first-order calculus in $\HH$ has closed form. Second,
given particles $X=\{x_i\}_{i=1}^N$, the finite-dimensional span
\begin{equation}
\label{eq:kernel_span}
\delta(\cdot)\;=\;\sum_{j=1}^{N} k(\cdot,x_j)\,c_j,
\qquad c_j\in\R^d,
\end{equation}
is the natural search space for ensemble updates: by the representer
argument, the optimizer of any regularized functional that touches the
particles only through $\{\delta(x_i)\}$ lies in this span. Writing
$K\in\R^{N\times N}$, with $K_{ij}=k(x_i,x_j)$ for the Gram matrix and stacking
coefficients row-wise into $C\in\R^{N\times d}$, the field values at the
particles are $D = KC$ and the RKHS norm is
$\norm{\delta}_{\HH}^2=\operatorname{tr}(C^\top K C)$.

A vector field $\delta$ induces the transport map $T_{\varepsilon}(x) = x +
\varepsilon\,\delta(x)$, which carries an ensemble with density $\rho$ to a
new density $\rho_\varepsilon$. For small $\varepsilon$, where
$T_\varepsilon$ is invertible, the transported density satisfies
\begin{align}
\label{eq:pushforward}
\log \rho_\varepsilon\bigl(T_\varepsilon(x)\bigr)
= \log\rho(x) - \log\bigl|\det\bigl(I + \varepsilon\,\nabla
\delta(x)\bigr)\bigr|,
\end{align}
so the log-determinant of the transport Jacobian measures the local change
in ensemble volume, and hence in entropy. This term reappears twice in our
method: implicitly, in the repulsion force of the Stein update, and
explicitly, in the merit function of our ensemble line search
(Section~\ref{sec:method_globalization}).

\subsection{Stein Variational Inference as Ensemble Optimization}
\label{sec:prelim_svgd}
Let $\pi$ be a target density on $\R^d$. Sampling from $\pi$ can be phrased
as minimizing the KL divergence $\KL(\rho\,\|\,\pi)$ over the ensemble
density $\rho$. SVGD performs this minimization by steepest descent
restricted to $\HH$: among all function perturbations $T_\varepsilon = \mathrm{id} +
\varepsilon\delta$ with $\norm{\delta}_{\HH}\le 1$, the one that
instantaneously decreases the KL fastest is
\begin{equation}
\label{eq:svgd_field}
\delta^{\star}_{\rho}(\cdot) \;\propto\;
\mathbb{E}_{x\sim\rho}\bigl[\,k(x,\cdot)\,\nabla_x\log\pi(x)
\;+\;\nabla_x k(x,\cdot)\,\bigr],
\end{equation}
the kernelized Stein operator applied to $\rho$.
Instantiating this at the $N$ particles gives the practical update: for each
particle $x_i$,
\begin{align}
\label{eq:svgd_empirical}
\phi(x_i) = \underbrace{\frac{1}{N}\sum_{j=1}^N
k(x_j,x_i)\,\nabla\log\pi(x_j)}_{\text{driving}}
+
\underbrace{\frac{1}{N}\sum_{j=1}^N \nabla_{x_j} k(x_j,x_i)}_{\text{repulsion}},
\end{align}
where the first term transports probability mass toward high-density regions
and the second, the derivative of the volume term in
\eqref{eq:pushforward} through the kernel, pushes particles apart and is
what prevents the ensemble from collapsing onto a single mode. Two
refinements are standard. \emph{Second-order (Stein--Newton) methods}
precondition the field with a curvature model of $-\log\pi$, either by a
global Newton system in the span \eqref{eq:kernel_span} or by cheaper
block-diagonal approximations.
\emph{Matrix-valued kernels} replace $k(x,y)$ by
$\mathcal{K}(x,y)\in\R^{d\times d}$, which allows the geometry of the target
to enter the function space itself; we use this device
for equality constraints below. Finally, the balance between driving and
repulsion in \eqref{eq:svgd_empirical} is exactly a temperature: annealing
it trades exploration against exploitation.
The abstraction to retain is this: \emph{an SVGD-type algorithm is an
optimization method whose iterate is the whole ensemble, whose search
direction is a vector field in $\HH$ sampled at the particles, and whose
objective couples an averaged cost to an entropy regularizer,}
\begin{equation}
\label{eq:free_energy}
E(\rho) \;=\; \mathbb{E}_{x\sim\rho}\bigl[V(x)\bigr] \;-\;
\gamma\, H(\rho),
\end{equation}
where $H(\rho)=-\mathbb{E}_{\rho}[\log\rho]$ is the ensemble entropy and the
temperature $\gamma>0$ sets their trade-off: the driving term reduces the
average cost $\mathbb{E}_\rho[V]$, while the repulsion increases the entropy
$H$ and keeps the particles spread out. Every globalization decision in
Section~\ref{sec:method} (what the line search measures, what
``convergence'' means for an ensemble that never stops jittering) is made
against this ensemble objective, not against any single particle's cost.

\subsection{Equality Constraints as Tangency Conditions}
\label{sec:prelim_constrained}
Constraining a sampler is not the same as constraining an optimizer: what
must respect $\F$ is the entire ensemble, and it must do so along the whole
flow, not only at the solution. For the equality part
$\mathcal{M}=\{x : h(x)=0\}$ of the feasible set, keeping every particle on
$\mathcal{M}$ at all times is equivalent to requiring the transport field to
be \emph{tangent}: $J_h(x)\,\delta(x)=0$ on $\mathcal{M}$, where
$J_h=\partial h/\partial x$. Prior work realizes this in one of three ways:
reparameterizing the domain through a mirror map; splitting the field into
tangential and normal components and using the normal one only to decay the
constraint residual; or ignoring tangency during the Stein step and
projecting particles back afterwards.
The device we adopt is the \emph{projected matrix-valued kernel} (the
\emph{orthogonal kernel} of \cite{zhang2022sampling}), later used for trajectory optimization (TO) in
\cite{power2024constrained}. Let
\begin{equation}
\label{eq:tangent_projector}
P(x) \;=\; I - J_h(x)^\top\bigl(J_h(x)J_h(x)^\top + \epsilon
I\bigr)^{-1}J_h(x)
\end{equation}
be the (regularized) orthogonal projector onto the tangent space of
$\mathcal{M}$ at $x$. Then
\begin{equation}
\label{eq:projected_kernel}
\mathcal{K}_P(x,y) \;=\; P(x)\,k(x,y)\,P(y)
\end{equation}
is a positive-semidefinite matrix-valued kernel whose RKHS contains only
fields that are pointwise tangent at the particles. Substituting
$\mathcal{K}_P$ for $k$ in \eqref{eq:svgd_field} projects the score and the
repulsion of each \emph{source} particle by its own $P(x_j)$ before the
kernel average, and the assembled field at each \emph{target} by $P(x_i)$
after it: the two projections do not commute with the sum and both are
required. 

As is conventional in constrained SVGD, the curvature term
$\nabla\!\cdot\!P$ of the manifold is dropped \cite{power2024constrained}. The
result is a diversity flow that slides \emph{along} the equality manifold:
it never spends its budget fighting the constraints, and it never undoes the
feasibility that the optimization layer (Section~\ref{sec:method})
establishes in the normal directions. Inequality constraints receive no such
reparameterization; their handling is precisely where our method departs from
prior work, through the SQP subproblem of
Section~\ref{sec:method_subproblem} rather than the squared slack variables
of \cite{power2024constrained}, which introduce spurious local optima near
active boundaries.
\section{Method: SteinSQP}
\label{sec:method}

Handling nonlinear constraints in constrained SVGD via sequential quadratic programming was first introduced in \cite{li2025constrained} and later analyzed in detail from the perspective of safe particle flows in \cite{yi2025constrained}. It keeps the two useful ingredients of SVGD, descent toward low-cost
motions and repulsion between particles, but does not rely on an exact
constrained Stein operator. Instead, every Stein-like proposal is made safe by
an SQP subproblem and by a line search that re-evaluates the nonlinear
constraints before accepting the step.

Let $X=\{x_i\}_{i=1}^N$ be the current ensemble, with
$x_i\in\R^d$. The objective is
$V(x)=\tfrac12\norm{r(x)}^2$, the score direction is
$s_i=-\nabla V(x_i)$, and the Gauss--Newton curvature is
$G_i=J_r(x_i)^\top J_r(x_i)$. Equality and inequality constraints are written
as $h(x)=0$ and $g(x)\le0$, with Jacobians $J_i^h$ and $J_i^g$ at particle
$i$. One SteinSQP iteration has four steps:
build a kernelized Stein descent model, solve a constrained quadratic
subproblem in kernel space, globalize the resulting ensemble transport, and
restart particles that remain in infeasible basins.

\subsection{A Stein-Like Descent Model}
\label{sec:method_model}

\subsubsection{Particle field}
For a scalar kernel $k$ with Gram matrix $K_{ij}=k(x_i,x_j)$, the practical
SVGD field at particle $i$ is split into a driving term and a repulsion term,
\begin{align}
\label{eq:stein_forces}
b_i &= \frac1N\sum_{j=1}^N K_{ji}\,s_j, \notag\\
\rho_i &= \frac1N\sum_{j=1}^N \nabla_{x_j} k(x_j,x_i).
\end{align}
The direction used by SteinSQP is
\begin{equation}
\label{eq:stein_model_gradient}
u_i = b_i+\gamma\rho_i ,
\end{equation}
where the temperature parameter $\gamma$ controls the strength of diversity. The first term pulls the
ensemble toward lower objective values; the second prevents collapse to a
single local solution. We use this field as a computational descent direction
for particles, not as an exact constrained sampling operator.

For equality constraints, an optional local tangent projection can be applied:
the projector $P_i=P(x_i)$ of \eqref{eq:tangent_projector} replaces $s_j$ and
$\nabla_{x_j}k$ in \eqref{eq:stein_forces} by their projected versions and
projects the final field, recovering the orthogonal SVGD of
\cite{power2024constrained} (Section~\ref{sec:prelim_constrained}). This is
only a local safeguard that biases the flow along the equality tangent space;
nonlinear feasibility is still enforced by the SQP and line-search layers
below.

\subsubsection{Curvature model}
Following the Stein Variational Newton (SVN) method~\cite{detommaso2018stein},
the Stein-like field is preconditioned by a curvature model of $-\log\pi$;
we use its cheaper block-Jacobi approximation. For each particle,
\begin{align}
\label{eq:stein_curvature}
H_i &=
\frac1N\sum_{j=1}^N K_{ji}^2 G_j  \nonumber \\
&+\frac{\gamma}{N}\sum_{j=1}^N
\nabla_{x_j}k(x_j,x_i)\nabla_{x_j}k(x_j,x_i)^\top
+\lambda I .
\end{align}
If tangent projection is used, the Gauss--Newton blocks and kernel gradients
are projected before entering \eqref{eq:stein_curvature}. The damping
$\lambda I$ keeps the proximal systems positive definite. This approximation
keeps only per-particle blocks, so the method avoids a coupled
$Nd\times Nd$ Newton system while still coupling particles through $K$.

\subsubsection{Trajectory kernels}
\label{sec:method_kernels}
For low-dimensional problems we use an RBF kernel. For trajectory variables,
we use an RBF under a fixed acceleration metric,
\begin{align}
\label{eq:structured_rbf_method}
k(x,y)&=\exp\!\left(
-\frac{(x-y)^\top M(x-y)}{2\hbar}\right), \notag\\
M&=\alpha I+\beta D_2^\top D_2 ,
\end{align}
where $D_2$ is the second-difference operator over trajectory knots and
$\hbar>0$ is the kernel bandwidth. This kernel measures similarity between
whole trajectories under an acceleration metric rather than the Euclidean
distance between stacked coordinates: through $D_2^\top D_2$, it compares
neighboring knots and so penalizes differences in curvature, treating two
motions as similar when they bend alike rather than merely passing through
nearby points. The identity term is important: without it, offsets and ramps
lie in the null space of $D_2^\top D_2$ and the kernel becomes insensitive to
some motion modes.

\subsection{Constrained Kernel-Space Step}
\label{sec:method_subproblem}
SteinSQP searches for an update field in the kernel span. Let
$C=(c_1,\ldots,c_N)^\top\in\R^{N\times d}$ be the coefficients and define the
particle update direction
\begin{equation}
\label{eq:kernel_direction_method}
d_i=(\mathcal K C)_i=\sum_{j=1}^N K_{ij}c_j ,
\end{equation}
with the projected-kernel version used when equality projection is enabled.
The coefficients $C$ are chosen by the SQP subproblem
\begin{equation}
\label{eq:stein_sqp_qp}
\begin{aligned}
\min_C\quad
&\sum_{i=1}^N
\left(\tfrac12 c_i^\top H_i c_i-u_i^\top c_i\right)\\
\mathrm{s.t.}\quad
&h_i+\varepsilon J_i^h d_i=0,\\
&g_i+\varepsilon J_i^g d_i\le0,
\qquad i=1,\ldots,N .
\end{aligned}
\end{equation}
The constraints are imposed on the particle displacement $d_i$, not on the
coefficient $c_i$. This matters because particles interact through the kernel:
the same coupling that spreads the ensemble also appears in the linearized
constraints. Equalities and inequalities are kept separate. Writing an
equality as two inequalities would create linearly dependent active rows at
every feasible point and unstable penalty multipliers.

\subsection{Inner Solve without Forming the KKT System}
\label{sec:method_pdhg}
At each outer iteration the SteinSQP direction is the minimizer of the quadratic
subproblem \eqref{eq:stein_sqp_qp}, posed not in the particle positions but in the
\emph{RKHS coefficients} $C=(c_1,\dots,c_N)$ that generate the transport field
$\phi(x_i)=\sum_j k(x_i,x_j)\,c_j$: the joint displacement of all particles is
$\varepsilon\,\mathcal K C$, where $\mathcal K$ is the kernel Gram operator,
$c_i\in\mathbb R^d$ the coefficient block of particle $i$, and $\varepsilon$ the SQP
step scale. The subproblem minimizes the block-separable model
$\sum_i\!\bigl(\tfrac12 c_i^\top H_i c_i-u_i^\top c_i\bigr)$, in which $u_i$ is the
Stein variational gradient of particle $i$ (its downhill score-driving pull plus the
inter-particle repulsion) and $H_i\succ0$ the corresponding block-Jacobi curvature
model of \eqref{eq:stein_curvature}, subject to the constraints \emph{linearized}
about the current particles, $h+B_h C=0$ and $g+B_g C\le0$. Here $h,g$ are the
equality/inequality residuals, $J^h,J^g$ their Jacobians, and
\begin{equation}
B_h C=\varepsilon J^h\mathcal K C,\qquad
B_g C=\varepsilon J^g\mathcal K C
\end{equation}
push a coefficient perturbation through the kernel into constraint space; we stack
them as $B=\bigl[\begin{smallmatrix}B_h\\ B_g\end{smallmatrix}\bigr]$.

Because the kernel couples all $N$ particles, forming and factorizing the dense KKT
system of this QP is prohibitive on a GPU. Even with a block-diagonal approximation, the Gram matrix still couples the per-particle constraints, preventing the subproblem from decomposing into independent particle-wise solves. We therefore dualize the constraints and
solve the resulting convex--concave saddle problem with the primal--dual scheme of
Chambolle and Pock~\cite{chambolle2011first, goldstein2013adaptive}, which needs only
\emph{applications} of $B$ and $B^\top$ and leaves the primal step separable.
Introducing an equality dual $\lambda$ and a \emph{nonnegative} inequality dual
$\mu\ge0$ (its sign enforces the KKT condition of $g\le0$), an extrapolated
coefficient $\bar C$, and primal and dual step sizes $\tau,\sigma>0$, one inner
iteration is
\begin{subequations}
\label{eq:pdhg_method}
\begin{align}
\lambda^{+} &= \lambda+\sigma(B_h\bar C+h), \label{eq:pdhg_lam}\\
\mu^{+} &= \bigl[\mu+\sigma(B_g\bar C+g)\bigr]_+, \label{eq:pdhg_mu}\\
\tilde C &= C-\tau B^\top
\begin{bmatrix}\lambda^{+}\\ \mu^{+}\end{bmatrix}, \label{eq:pdhg_predict}\\
c_i^{+} &= (I+\tau H_i)^{-1}(\tilde c_i+\tau u_i), \label{eq:pdhg_prox}\\
\bar C^{+} &= C^{+}+\theta(C^{+}-C). \label{eq:pdhg_extrap}
\end{align}
\end{subequations}
Steps \eqref{eq:pdhg_lam}--\eqref{eq:pdhg_mu} take a gradient \emph{ascent} of the
duals toward feasibility of the linearized constraints, with $\sigma$ the dual step;
the clamp $[\,\cdot\,]_+$ is the proximal operator of the constraint $\mu\ge0$, i.e.
a projection onto the nonnegative orthant. Steps
\eqref{eq:pdhg_predict}--\eqref{eq:pdhg_prox} form the primal gradient step
$\tilde C$ (with primal step $\tau$) and apply the proximal operator of the objective
quadratic; since $H=\operatorname{diag}(H_i)$ is block diagonal, this prox
\emph{decouples} into $N$ independent $d\times d$ SPD solves, one per particle, done
in parallel. Because $H_i$ and $\tau$ are fixed across the inner loop, each
$(I+\tau H_i)^{-1}$ is factorized once and reused for every inner iteration, so the
prox is cheap. Step \eqref{eq:pdhg_extrap} is the over-relaxation that gives the
scheme its accelerated, provably convergent rate. The step sizes must satisfy the
PDHG stability bound $\tau\sigma\lVert B\rVert^2\le1$: we fix $\tau$ from a bound on
$\lVert H\rVert$ (so each prox is well conditioned) and take $\sigma$ to saturate the
bound up to a safety factor, estimating the coupling norm $\lVert B\rVert$ by a few
power-iteration sweeps of $B^\top B$, never assembling $B$, only applying
$\mathcal K,\,J^h,\,J^g$ and their adjoints. Thus no KKT matrix is ever formed or
factorized: each of the fixed number of inner iterations costs a handful of
matrix-free operator applications plus $N$ tiny per-particle solves, giving $O(Nd)$
memory and an inner solve that parallelizes across particles. This is the root reason SteinSQP is efficient on a GPU: PDHG needs only dense kernel and Jacobian products and per-particle solves, exactly the regular, batched operations GPUs are built for.

We did explore exploiting the sparsity of the constraint Jacobians instead, but
found it does not pay off: in our tests a sparse QP solver ran at least an order of magnitude or more slower on GPU, as sparse factorization maps poorly onto the hardware and
its irregular structure outweighs the benefit of the reduced flop count. A dense
matrix-free PDHG sweep, by contrast, keeps every operation regular and fully
parallel, so we use it rather than a sparse QP solver for this subproblem.

\subsection{Globalization}
\label{sec:method_globalization}
The SQP subproblem only sees a linearized model. A step is accepted only after
testing the nonlinear objective and constraints. Let
$\Delta_i=\varepsilon d_i$ and let $\Phi_i=\partial\Delta_i/\partial x_i$ be
the Jacobian of the particle transport. For a candidate step length $\alpha$,
the line search evaluates
\begin{equation}
\label{eq:corrected_candidate}
y_i(\alpha)=x_i+\alpha\Delta_i+s_i(\alpha),
\end{equation}
where $s_i(\alpha)$ is a second-order correction defined below. The accepted
step is selected by the ensemble merit
\begin{align}
\label{eq:ensemble_merit}
M(\alpha)={}&
\frac1N\sum_i V(y_i(\alpha))
-\gamma\frac1N\sum_i\log\det(I+\alpha\Phi_i) \notag\\
&+\nu\frac1N\sum_i
\left(\norm{h(y_i(\alpha))}_1+
\norm{g^+(y_i(\alpha))}_1\right).
\end{align}
The log-determinant term is the same local volume effect that produces Stein
repulsion. Including it prevents the line search from always preferring
collapse toward the lowest-cost particle. The line search globalizes on this
nonsmooth $\ell_1$ exact-penalty merit, whose constraint term, the mean
per-particle violation, enters weighted by the penalty parameter $\nu$.
Exact-penalty theory fixes what this weight must be: the SQP step is a descent
direction for the merit, and a constrained minimizer is one of its stationary
points, precisely when $\nu$ dominates the optimal multipliers,
$\nu>\lVert(\lambda^\star,\mu^\star)\rVert_\infty$, with the $\ell_\infty$ norm
dual to the $\ell_1$ violation measure~\cite{nocedal2006numerical}. A single
hand-tuned $\nu$ cannot meet this robustly: too small and the penalty loses
exactness, so the search stalls at infeasible points; too large and the merit
is swamped by feasibility, forcing vanishing steps and ill-conditioning. We
instead read the multiplier estimates straight off the inner solve, since the
PDHG duals $\lambda,\mu$ are exactly the current estimates of
$\lambda^\star,\mu^\star$, and set
\begin{equation}
\label{eq:exact_penalty_method}
\nu=\max\!\left(\nu_{\min},\;
\rho_\nu\max\bigl(\norm{\lambda}_\infty,\norm{\mu}_\infty\bigr)\right),
\end{equation}
with a safety factor $\rho_\nu>1$ that keeps $\nu$ strictly above the live
multipliers and a floor $\nu_{\min}$ that guards the fully feasible case
$\lambda,\mu\to0$. The weight is thus \emph{just} large enough to stay exact
but no larger, and it tracks the multipliers automatically, rising as they grow
near strongly active constraints and relaxing as they settle toward the
solution, so feasibility is never balanced by a manually fixed penalty scale.

\subsubsection{Second-order correction (SOC)}
To reduce the gap between the linearized constraints and the true nonlinear
constraints, every line-search candidate is retracted once by
\begin{equation}
\label{eq:soc_method}
s_i(\alpha)=-
\bigl(\bar J_i^\top\bar J_i+\eta_i I\bigr)^{-1}
\bar J_i^\top
\bar c_i(x_i+\alpha\Delta_i).
\end{equation}
Here $\bar c_i$ stacks the equality residuals and the currently violated
inequality rows, and $\bar J_i$ is the corresponding block of the current
constraint Jacobian. The correction is computed at the candidate point rather
than only at the full step, because backtracking can change which inequalities
are violated. This is a standard SQP trick to overcome the Maratos
effect~\cite{nocedal2006numerical}. In our experiments, without SOC SteinSQP
diverges on difficult NLP problems.

\subsubsection{Line search}
The step length is shared by the whole ensemble. SteinSQP evaluates a fixed
geometric ladder of candidates in parallel and accepts the largest one that
satisfies an Armijo decrease test on \eqref{eq:ensemble_merit}. If no candidate
passes, a filter rule may accept a step that strictly reduces constraint
violation without increasing the transport objective too much. Candidates with
non-invertible transport, detected by the log-determinant term, are rejected.

\subsection{Escaping Infeasible Basins}
\label{sec:method_resampling}
Nonconvex inequalities can trap a particle in an infeasible homotopy class. At
a fixed period, particles that still fail the task feasibility test are
restarted from feasible particles in the same ensemble:
\begin{align}
\label{eq:resampling_method}
x_i &\leftarrow x_{j(i)} + P_{j(i)}\xi_i,\notag\\
j(i)&\sim\operatorname{Unif}\{j:x_j\ \mathrm{feasible}\},\qquad
\xi_i\sim\mathcal N(0,\sigma_r^2I).
\end{align}
The projection is used when a tangent projector is available; otherwise the
restart is an ordinary local perturbation. This mechanism is not meant to
replace constraint handling. It only gives particles that are locally stuck
another feasible basin from which the Stein repulsion can again diversify them.

\subsection{Stopping and Computation}
\label{sec:method_gpu}
The solver stops when all particles are feasible and the ensemble merit level
has plateaued over a resampling period. We do not stop on small particle
motion, because repulsion can keep a well-converged ensemble moving slightly.
All loops have fixed shapes: PDHG uses a fixed iteration count, the line search
uses a fixed candidate ladder, and resampling is implemented by masked updates.
The dense backend costs $O(N^2d)$ for kernel contractions, $O(Nd^3)$ for the
batched proximal solves, and $O(L(N^2d+N(m_h+m_g)d))$ for $L$ PDHG iterations.
For trajectory kernels, multiplication by $M$ in
\eqref{eq:structured_rbf_method} uses block-banded storage, while the ensemble
coupling remains a dense $N\times N$ particle operation. 

The full SteinSQP algorithm is given in Algorithm~\ref{alg:steinsqp}.

\begin{algorithm}[t]
\caption{SteinSQP}
\label{alg:steinsqp}
\begin{algorithmic}[1]
\Require initial ensemble $X^0$, kernel $k$, step size $\varepsilon$
\For{$r=0,\ldots,r_{\max}-1$}
    \State evaluate $V,h,g,J^h,J^g$ at the current ensemble
    \State build the Stein-like model $u_i,H_i$ \Comment{\eqref{eq:stein_model_gradient}, \eqref{eq:stein_curvature}}
    \State solve the constrained kernel-space QP \eqref{eq:stein_sqp_qp} by PDHG \eqref{eq:pdhg_method}
    \State set the exact-penalty weight $\nu$ from the QP duals \Comment{\eqref{eq:exact_penalty_method}}
    \State accept a corrected ensemble step by the merit \eqref{eq:ensemble_merit}
    \If{a resampling period is reached}
        \State restart particles that remain infeasible \Comment{\eqref{eq:resampling_method}}
        \State \textbf{stop} if all particles are feasible and the merit has plateaued
    \EndIf
\EndFor
\Ensure feasible, diverse particle ensemble
\end{algorithmic}
\end{algorithm}
\section{Experiments}
\label{sec:experiments}

The experiments are organized around three questions. First, does SteinSQP
return a feasible \emph{ensemble}, not only a feasible best particle? Second,
does the globalized Stein--Newton flow reach that ensemble faster than
first-order constrained Stein optimization and than serial multistart
nonlinear programming? Third, when a task admits several solutions, does the
ensemble retain them instead of collapsing onto a single mode?

All experiments are conducted on a desktop workstation with an AMD
Ryzen~9~7950X3D CPU and an NVIDIA GeForce RTX~4080 GPU (16\,GB): the
JAX-based SteinSQP and CSVTO run on the GPU, while IPOPT runs on the CPU.

\begin{figure*}[htbp]
\centering
\includegraphics[width=0.49\textwidth]{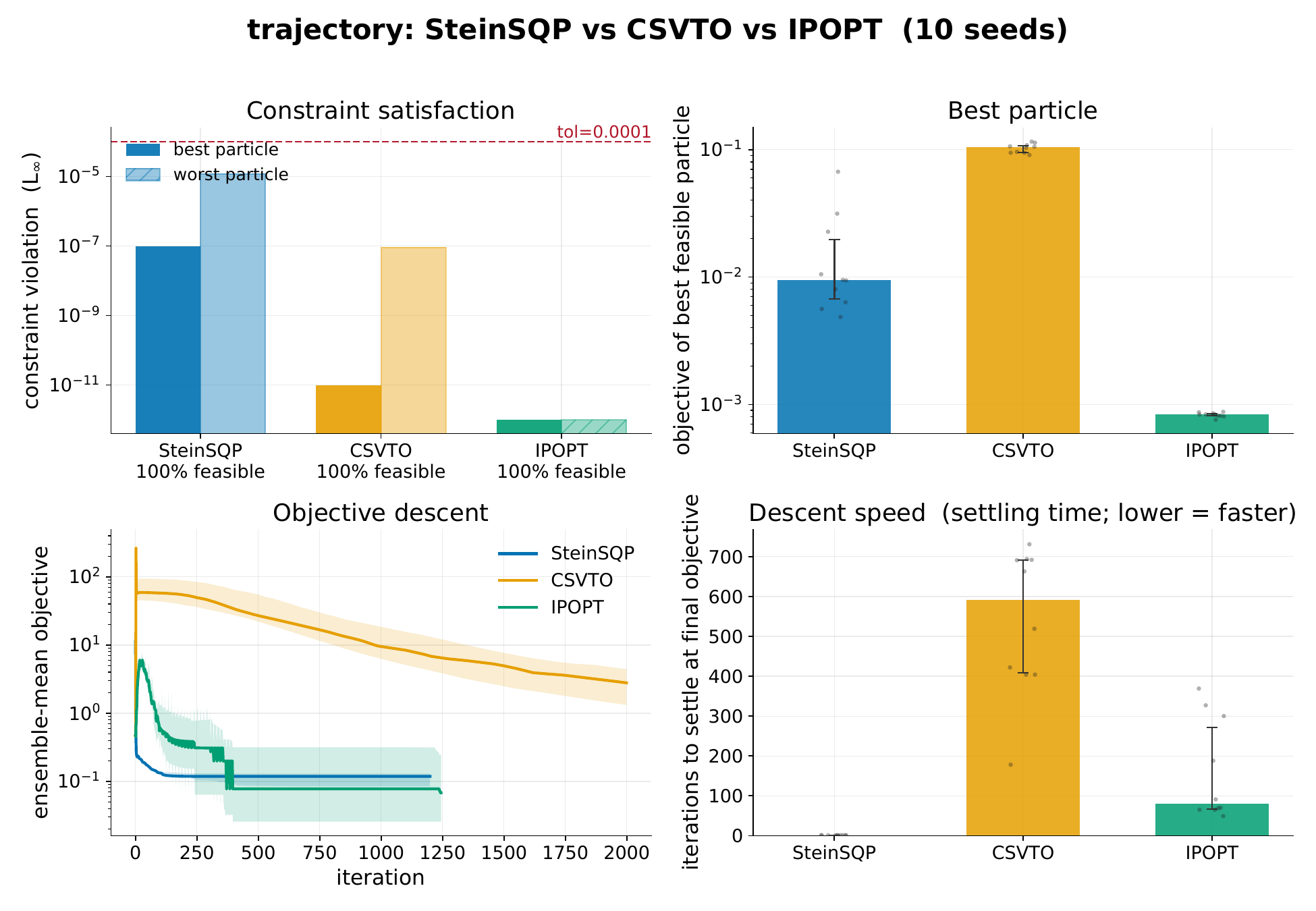}\hfill
\includegraphics[width=0.49\textwidth]{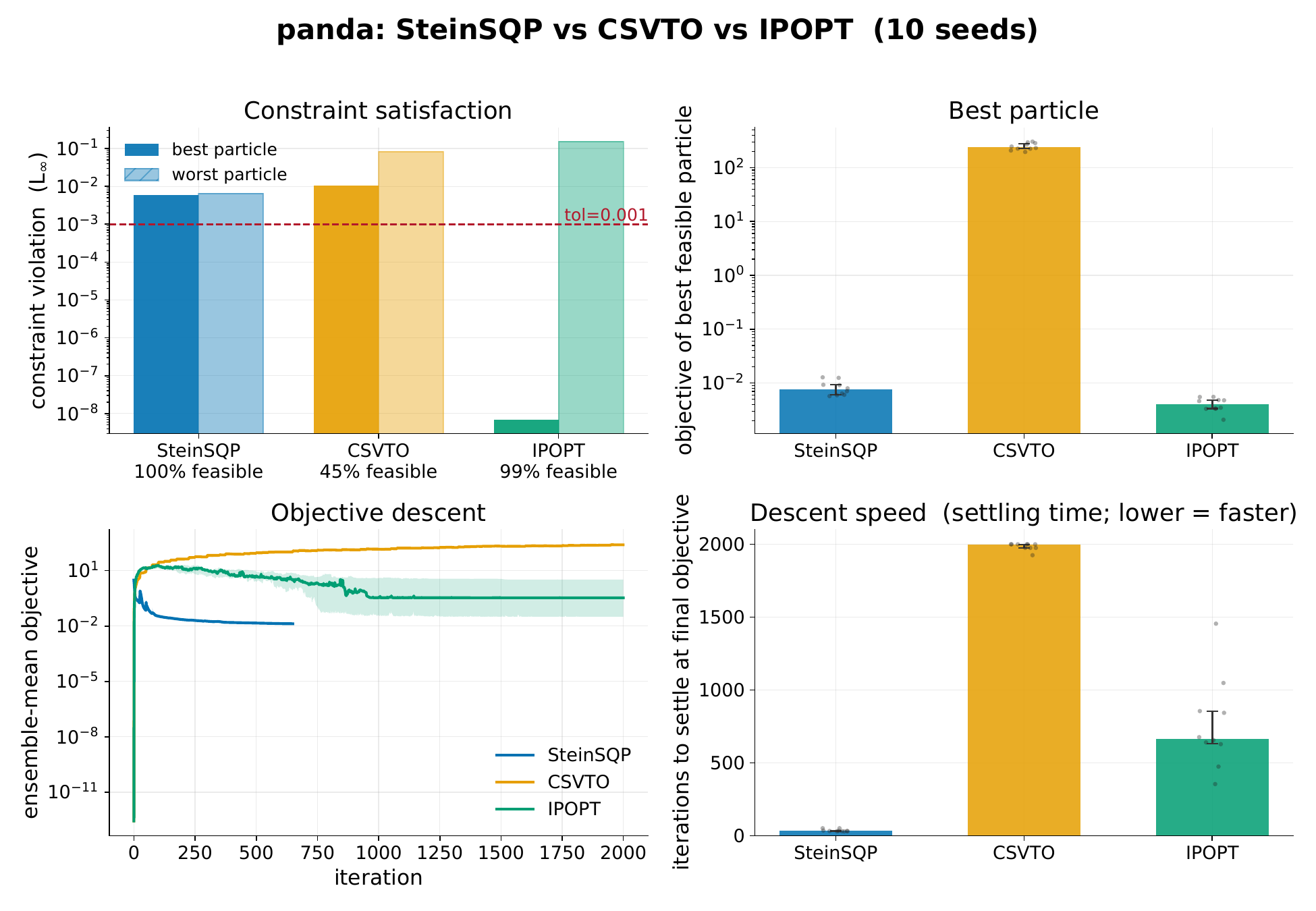}
\caption{Convergence diagnostics on trajectory (left) and panda (right); all
statistics are medians with IQR over $10$ seeds. Per task: best- and
worst-particle $L_\infty$ violation against the task tolerance with the
feasible fraction, best feasible objective, ensemble-mean objective per
iteration, and the settling time, the iteration after which the mean
objective stays within a $10\%$ band of its final plateau, so oscillation
counts in full.}
\label{fig:metrics}
\end{figure*}

\subsection{Benchmarks, Baselines, and Protocol}
\label{sec:experiments_setup}

\textbf{Tasks.}
We evaluate on five tasks ordered from diagnostic to robot-scale, with the
ensemble size $N$ fixed per task.
\emph{Annulus} ($N{=}88$): sampling a 2-D Gaussian restricted to the band
$2.5\le\|x\|\le 3$; the target density is known in closed form, so
distributional behavior can be judged against ground truth.
\emph{Trajectory} ($N{=}16$): planar spline trajectory optimization (33
control points) through scattered circular obstacles treated as nonlinear
collision inequalities.
\emph{Surface} ($N{=}32$): traversing a curved terrain while avoiding holes;
keeping the path on the terrain contributes a large system of equality
constraints on top of the clearance inequalities.
\emph{Panda throw} ($N{=}16$): a 7-DoF manipulator throws a ball into a
basket, under joint limits, arm and ball collision clearance, a pinned start
pose, and a ballistic landing constraint (\Cref{fig:panda_throw}).
\emph{Push-box} ($N{=}16$): a pusher must bring a box to a goal pose through a
smoothed single-shooting contact rollout; distinct contact faces and rotation
directions make the task multimodal by construction. We defer the detailed NLP problem definitions to \Cref{app:nlp_problems}.

\textbf{Baselines.}
The baselines separate the two roles of the method. CSVTO, a first-order
constrained Stein trajectory optimizer sharing our kernels, tests whether the
second-order Stein geometry and the ensemble globalization are necessary.
Multistart IPOPT, launched once from every particle of the same initial
ensemble, tests whether independent deterministic solves can replace a coupled
particle flow. SteinSQP and CSVTO run as batched float32 ensemble solvers on
the GPU; IPOPT solves each start serially in float64 (its Python interface
does not support float32), with exact sparse Lagrangian Hessians on the three
low-dimensional tasks and L-BFGS on panda and push-box.

\textbf{Protocol.}
All results aggregate $10$ seeds per (task, method) pair. For seed $s$ all
three methods receive the identical initial ensemble, so IPOPT's starts are
bit-identical to the particle methods' particles. Each method keeps its
hand-tuned per-task configuration and runs under a shared stopping rule: a run
ends as soon as \emph{all} particles are feasible at the task tolerance and
the ensemble merit has plateaued, with a hard cap of $2000$ iterations.
CSVTO has no globalization and never reaches this test, so it consumes the
full budget on every task. IPOPT runs at KKT tolerance $10^{-4}$, with its
constraint-violation tolerance pinned to the same task feasibility tolerance
the particle methods must reach. Every final ensemble is scored by one
scorer: the same per-particle objective, the same $L_\infty$ violation, and
the same feasibility predicate at the true task tolerance.
All timings are \emph{warm}: an untimed pass first triggers XLA compilation
(and compiles IPOPT's autodiff callbacks), so the reported times measure the
solve, not compilation.

\subsection{Metrics}
\label{sec:experiments_metrics}

The first primary metric is \emph{particle-wise feasibility}: the fraction of
all returned particles, across all seeds, that satisfy the task constraints.
This is the success criterion appropriate for a sampler, since a method that
returns one feasible trajectory and many unusable particles has not solved the
ensemble problem. The second primary metric is \emph{warm time-to-solution}:
for SteinSQP and CSVTO the wall time of the single batched ensemble solve, for
IPOPT the sum of its $N$ serial multistart solves. Since the Python wrapper
adds callback overhead to IPOPT, we also report its callback-free internal
time (wall time minus bracketed callback time) as a lower bound on a native
implementation.

Secondary metrics describe sampling quality. On the annulus we report the
unbiased squared maximum mean discrepancy (MMD$^2$) between the final ensemble
and $4000$ rejection samples of the true truncated Gaussian, using an RBF
kernel with a \emph{fixed} bandwidth of $0.5$ set from the band geometry (a
data-adaptive bandwidth would differ per ensemble and make the scores
incomparable across methods). On push-box we report \emph{mode coverage}, the
number of distinct (rotation sign, contact face) labels among feasible
particles, and the \emph{feasible spread}, the mean pairwise distance between
feasible particles. We also track the best feasible objective, the median
over seeds of the lowest cost among feasible particles, as a diagnostic
rather than a claim: IPOPT is expected to be strong at single-solution local
optimization, while SteinSQP targets feasible, diverse ensembles.

\subsection{Main Results}
\label{sec:experiments_results}

\Cref{tab:experiment_main} gives the headline results. SteinSQP is the only
method that returns a fully feasible ensemble on every task: $1680/1680$
particles over all seeds, against $1504/1680$ for CSVTO and $1599/1680$ for
IPOPT. Every method finds at least one feasible particle in every seed; the
gap opens at the ensemble level, which is exactly where a sampler is judged.
SteinSQP also triggers the shared stopping test on every task (median
$238$--$1080$ of the $2000$ allotted iterations), so its times reflect
convergence, not budget exhaustion.

\begin{table}[hbp]
\centering
\caption{Primary results, median over $10$ seeds per task. \emph{Success}:
particle-wise feasibility (feasible / returned particles). \emph{Wall time}:
warm time-to-solution, one batched solve for SteinSQP and CSVTO, the serial sum
over $N$ multistart solves for IPOPT. \emph{Iters.}: iterations at termination
(cap $2000$). $^{\dagger}$CSVTO never reaches the stopping test and exhausts
the budget. $^{\ddagger}$Interior-point iterations per start, not
cost-comparable per iteration.}
\label{tab:experiment_main}
\begin{tabular}{llrrr}
\toprule
Task & Method & Success & Wall time (s) & Iters. \\
\midrule
\multirow{3}{*}{\begin{tabular}[c]{@{}l@{}}Annulus\\($N{=}88$)\end{tabular}}
& SteinSQP & 100\% (880/880) & 0.344 & 238 \\
& CSVTO    & 100\% (880/880) & 0.192 & 2000$^{\dagger}$ \\
& IPOPT    & 100\% (880/880) & 1.247 & 9$^{\ddagger}$ \\
\midrule
\multirow{3}{*}{\begin{tabular}[c]{@{}l@{}}Trajectory\\($N{=}16$)\end{tabular}}
& SteinSQP & 100\% (160/160) & 0.554 & 462 \\
& CSVTO    & 100\% (160/160) & 16.791 & 2000$^{\dagger}$ \\
& IPOPT    & 100\% (160/160) & 12.679 & 146$^{\ddagger}$ \\
\midrule
\multirow{3}{*}{\begin{tabular}[c]{@{}l@{}}Surface\\($N{=}32$)\end{tabular}}
& SteinSQP & 100\% (320/320) & 0.962 & 1080 \\
& CSVTO    & 100\% (320/320) & 7.018 & 2000$^{\dagger}$ \\
& IPOPT    & 100\% (320/320) & 4.411 & 34$^{\ddagger}$ \\
\midrule
\multirow{3}{*}{\begin{tabular}[c]{@{}l@{}}Panda\\($N{=}16$)\end{tabular}}
& SteinSQP & 100\% (160/160) & 2.613 & 488 \\
& CSVTO    & 45\% (72/160)   & 52.789 & 2000$^{\dagger}$ \\
& IPOPT    & 99\% (158/160)  & 282.625 & 895$^{\ddagger}$ \\
\midrule
\multirow{3}{*}{\begin{tabular}[c]{@{}l@{}}Push-box\\($N{=}16$)\end{tabular}}
& SteinSQP & 100\% (160/160) & 1.917 & 562 \\
& CSVTO    & 45\% (72/160)   & 3.385 & 2000$^{\dagger}$ \\
& IPOPT    & 51\% (81/160)   & 53.763 & 312$^{\ddagger}$ \\
\bottomrule
\end{tabular}
\end{table}

\textbf{Against first-order Stein (CSVTO).}
On the three geometric tasks CSVTO also reaches full feasibility, but because
it never satisfies the stopping test it pays the full budget, ending up
$30\times$ (trajectory) and $7\times$ (surface) slower than SteinSQP; only on
the 2-D annulus, where a first-order update is nearly free, is it faster
($0.19$\,s vs.\ $0.34$\,s). On the two robot-scale tasks its lack of
globalization becomes a correctness failure, not merely a speed one: it loses
more than half the ensemble ($45\%$ feasible on both panda and push-box), and
the cost of its surviving particles degrades by up to four orders of magnitude
(\Cref{tab:experiment_cost}). The second-order geometry and the ensemble line
search are therefore not refinements; they are what makes the constrained
Stein flow dependable.

\textbf{Against multistart IPOPT.}
IPOPT matches feasibility on the smooth tasks and nearly so on panda
($99\%$), but as a serial multistart it is $4.6$--$23\times$ slower than the
batched ensemble solve there, and $108\times$ slower on panda
($282.6$\,s vs.\ $2.6$\,s). This is not an artifact of wrapper overhead: even
its callback-free internal lower bound ($244.9$\,s on panda, $12.6$\,s on
push-box) exceeds SteinSQP's total wall time on every task. On the
contact-rich push-box rollout, interior-point iterations frequently stall, and
only $51\%$ of starts end feasible. A set of independent deterministic solves
thus reproduces neither the reliability nor the cost profile of the coupled
flow.

\textbf{Ensemble-level convergence.}
\Cref{fig:metrics} compares convergence from the particle-ensemble perspective. On both trajectory and panda, SteinSQP decreases the ensemble-mean objective more rapidly and reaches its final plateau earlier than the baselines, while also driving the worst-particle constraint violation below the task tolerance. This is the iteration-level convergence advantage reported in our results: the Stein--Newton flow makes faster progress per ensemble update, rather than merely obtaining a lower wall-clock time from batching. CSVTO, without globalization, exhibits larger oscillations and settles last or not within the iteration budget. IPOPT can reduce the objective quickly within individual local solves, but its progress is distributed across independent serial starts rather than a coupled ensemble update.

\begin{figure*}[t]
\centering
\includegraphics[width=\textwidth]{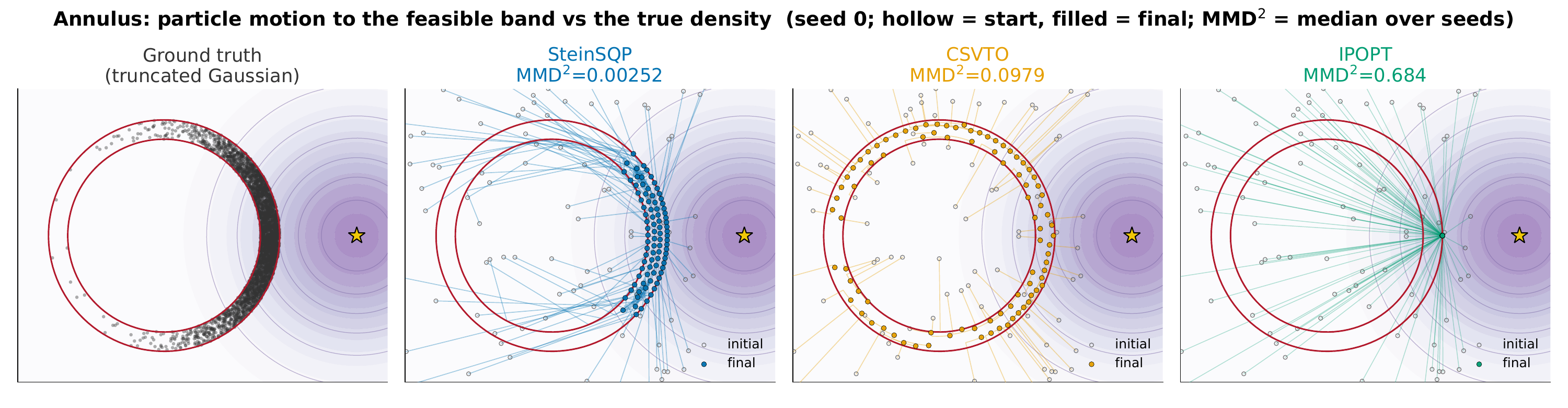}
\caption{Annulus particle motion vs.\ the known target. Left: $4000$
ground-truth samples of the Gaussian truncated to the feasible band. Right:
per-method motion of the shared seed-$0$ ensemble from initialization (hollow)
to the final particles (filled); the Stein methods show per-iteration trails,
while IPOPT, which exposes no per-iterate ensemble, is drawn as straight
initial-to-final segments. Panel titles report the median MMD$^2$ over
$10$ seeds.}
\label{fig:annulus_motion}
\end{figure*}
\subsection{Sampling Quality and Diversity}
\label{sec:experiments_diversity}

\begin{table}[hbp]
\centering
\caption{Best feasible objective: median over seeds of the lowest cost among
feasible particles. A diagnostic, not the headline: IPOPT is an exact float64
local optimizer and should win it wherever it converges (on push-box, among the
$51\%$ of starts that do).}
\label{tab:experiment_cost}
\begin{tabular}{lccc}
\toprule
Task & SteinSQP & CSVTO & IPOPT \\
\midrule
Annulus    & $0.500$ & $0.530$ & $0.500$ \\
Trajectory & $9.46\times10^{-3}$ & $1.04\times10^{-1}$ & $8.33\times10^{-4}$ \\
Surface    & $325.9$ & $473.5$ & $310.6$ \\
Panda      & $7.53\times10^{-3}$ & $2.39\times10^{2}$ & $4.06\times10^{-3}$ \\
Push-box   & $6.07\times10^{-2}$ & $1.50\times10^{-1}$ & $1.23\times10^{-7}$ \\
\bottomrule
\end{tabular}
\end{table}

\begin{table}[hbp]
\centering
\caption{Sampling diagnostics, median over seeds. MMD$^2$: unbiased RBF MMD
(fixed bandwidth $0.5$) between the final ensemble and $4000$ ground-truth
samples. Spread: mean pairwise distance among \emph{feasible} particles; on the
annulus it diagnoses collapse, on push-box larger means more distinct feasible
alternatives. Modes: distinct (rotation sign, contact face) labels among
feasible particles.}
\label{tab:experiment_secondary}
\setlength{\tabcolsep}{3pt}
\begin{tabular}{lccc}
\toprule
Metric & SteinSQP & CSVTO & IPOPT \\
\midrule
Annulus: MMD$^2$ $\downarrow$
& $\mathbf{2.52\times10^{-3}}$ & $9.79\times10^{-2}$ & $6.84\times10^{-1}$ \\
Annulus: spread
& $1.32$ & $3.48$ & $0.00$ \\
\midrule
Push-box: feasible modes $\uparrow$
& $2.0$ & $2.3$ & $2.0$ \\
Push-box: feasible spread $\uparrow$
& $\mathbf{3.36}$ & $2.37$ & $1.50$ \\
\bottomrule
\end{tabular}
\end{table}

\textbf{Distributional accuracy (annulus).}
The annulus is the one task with an exact target density, and it cleanly
separates the three methods (\Cref{tab:experiment_secondary},
\Cref{fig:annulus_motion}). SteinSQP attains an MMD$^2$ of
$2.52\times10^{-3}$, $39\times$ lower than CSVTO and $271\times$ lower than
IPOPT. \Cref{fig:annulus_motion} shows why: all $88$ IPOPT starts contract
onto the single constrained optimum (spread $0.00$), so a multistart optimizer
is not a sampler even when every start succeeds, while CSVTO keeps particles
spread but transports them so slowly along the heavily constrained band that
its ensemble never matches the target's radial and angular profile within the
budget. SteinSQP reproduces both marginals while keeping every particle in
the band.

\begin{figure}[t]
\centering
\includegraphics[width=\columnwidth]{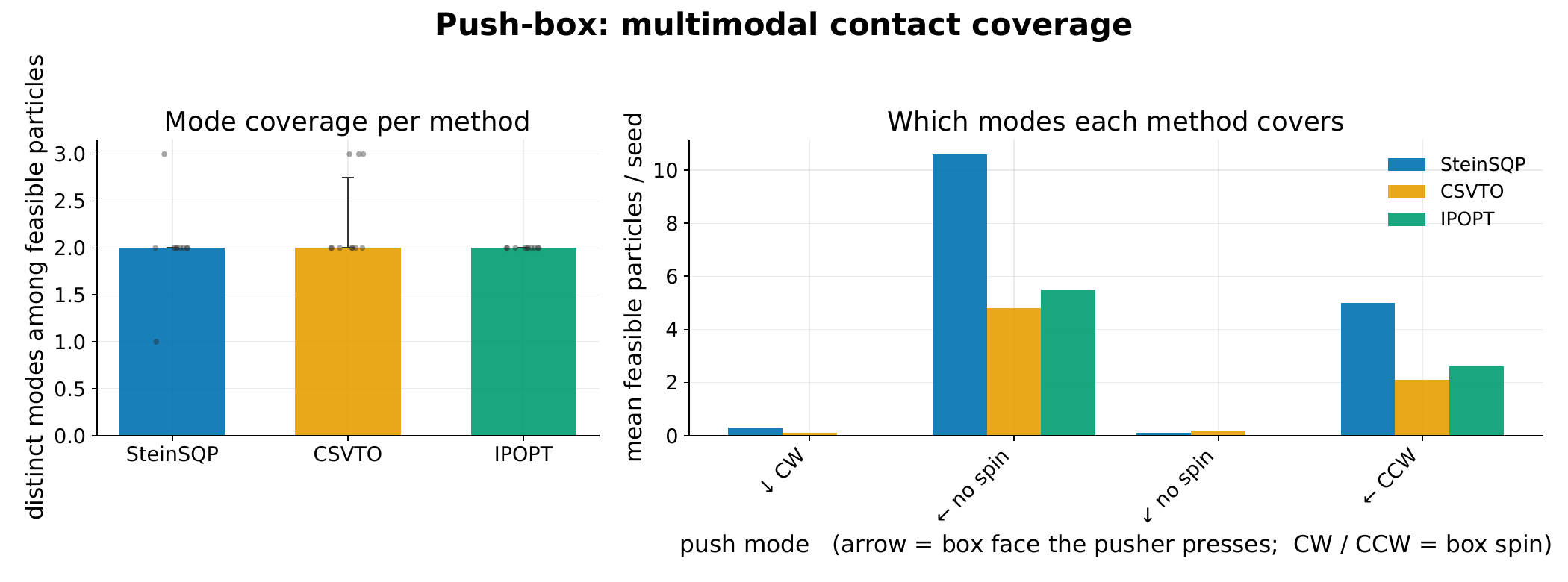}
\caption{Push-box mode coverage. Left: distinct contact modes among feasible
particles per seed (median, IQR). Right: per-mode occupancy, the mean feasible
particles per seed landing in each observed (spin, contact-face) mode.}
\label{fig:pushbox_modes}
\end{figure}

\begin{figure}[tbp]
\centering
\includegraphics[width=\columnwidth]{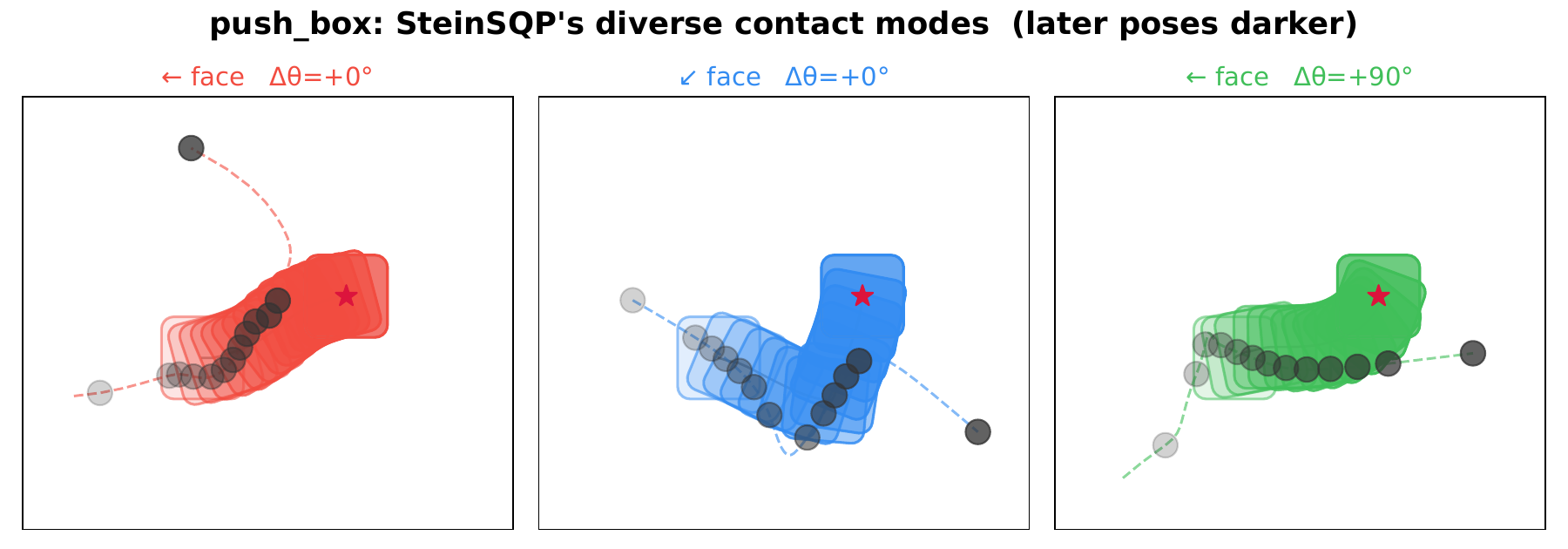}
\caption{Distinct feasible contact modes from a single SteinSQP solve, one
panel per mode: box poses are ghosted along the box's path (later poses
darker), the pusher path is dashed; the modes differ in contact face and net
rotation direction.}
\label{fig:pushbox_timelapse}
\end{figure}

\textbf{Multimodal coverage (push-box).}
Push-box is multimodal by construction, and here coverage must be read jointly
with feasibility (\Cref{tab:experiment_secondary},
\Cref{fig:pushbox_modes}). SteinSQP discovers as many feasible contact modes
as multistart IPOPT ($2.0$) while returning $100\%$ of its particles feasible
against IPOPT's $51\%$, and its feasible plans are the most mutually distinct
(spread $3.36$ vs.\ $2.37$ and $1.50$). CSVTO's nominally highest coverage
($2.3$) is obtained only after discarding $55\%$ of its ensemble.
\Cref{fig:pushbox_timelapse} renders the qualitatively different feasible
pushes, opposite rotation directions and contact faces, returned by a single
SteinSQP solve.

\begin{figure}[t]
\centering
\includegraphics[width=\columnwidth]{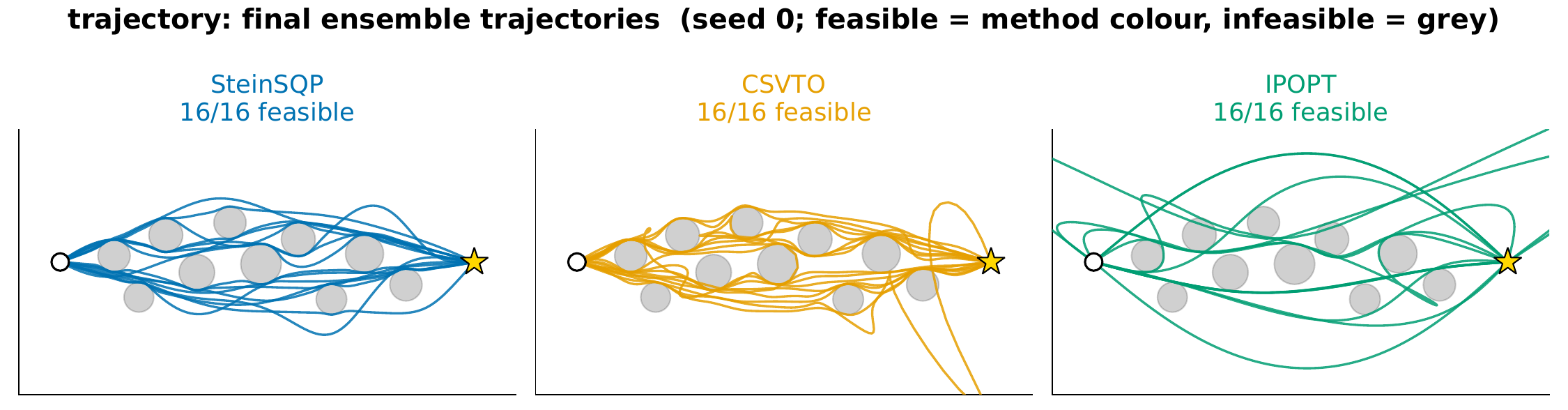}
\caption{Final ensembles on the trajectory task for one shared seed: all
methods start from the identical ensemble; feasible plans are drawn in the
method color, infeasible plans in grey (counts in the panel titles).}
\label{fig:final_traj_trajectory}
\end{figure}

\begin{figure}[t]
\centering
\includegraphics[width=\columnwidth]{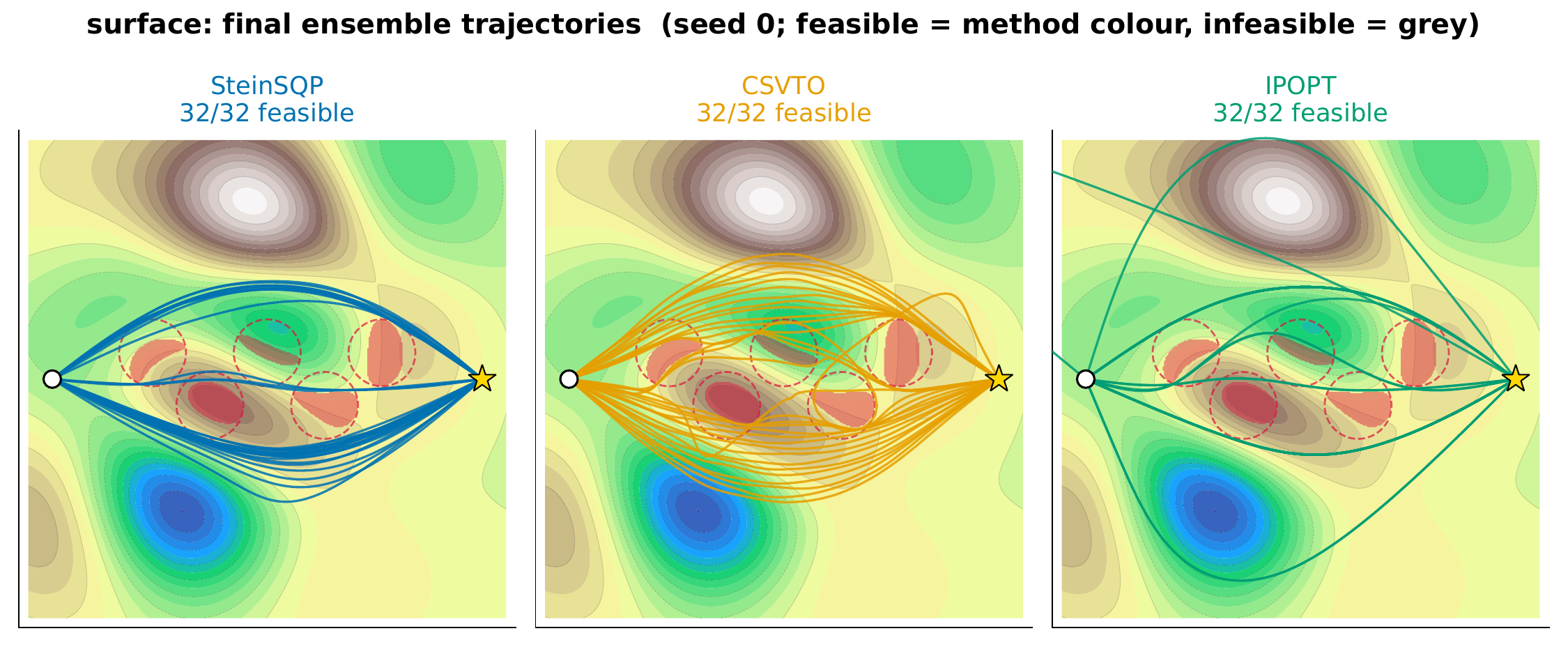}
\caption{Final ensembles on the surface task for one shared seed: all methods
start from the identical ensemble; feasible plans are drawn in the method
color, infeasible plans in grey (counts in the panel titles).}
\label{fig:final_traj_surface}
\end{figure}

\textbf{Where multistart keeps diversity.}
On trajectory, surface, and panda, IPOPT's independent starts do end at
distinct local optima, inheriting the spread of the shared initialization
(\Cref{fig:final_traj_trajectory,fig:final_traj_surface}); on such tasks the
advantage of the coupled flow is not diversity but cost, one batched solve
versus a serial sum (\Cref{tab:experiment_main}). The ensemble-level
advantages of the flow appear precisely where the paper claims them: where the
task is distributional (annulus) or contact-rich (push-box).

\textbf{Best feasible cost.}
\Cref{tab:experiment_cost} follows the expected pattern. IPOPT, an exact
float64 local optimizer, attains the lowest best-particle cost wherever it
converges; SteinSQP stays close on the robot-scale tasks (within $5\%$ on
surface, the same order of magnitude on panda) while returning the
\emph{entire} ensemble feasible, and CSVTO's best particle degrades by one to
four orders of magnitude on the hard tasks. This is the trade the method is
designed to make: a margin of best-particle optimality in exchange for a fully
usable, diverse set of plans.

\subsection{Takeaway}
\label{sec:experiments_takeaway}

The SQP layer and the ensemble-level globalization are what make the Stein
flow usable under nonlinear constraints. SteinSQP is the only method that
turns the whole ensemble into feasible motion output ($1680/1680$ particles),
and it reaches that ensemble at the lowest warm time-to-solution on every task
but the 2-D diagnostic, up to $108\times$ faster than serial multistart
IPOPT, whose internal solver time alone exceeds SteinSQP's total on all five
tasks. At the same time it preserves the reason to use a Stein method at all:
a close match to the target distribution where one exists ($39$--$271\times$
lower MMD$^2$) and several qualitatively distinct feasible modes in a single
solve. The two baselines exhibit exactly the failure modes the method is
built to remove: the first-order flow returns diverse but partially infeasible
ensembles on hard constraints, and multistart nonlinear programming pays a
serial cost and collapses distributionally where the task is sampling.
\section{Conclusion}
\label{sec:conclusion}

We presented SteinSQP, a constrained Stein variational inference method for sampling diverse feasible robot motions. Rather than relaxing nonlinear constraints through slack variables, penalties, or post-hoc projection, SteinSQP embeds linearized equality and inequality constraints directly into the kernel-coupled ensemble displacement, yielding an SQP-like constrained Stein--Newton update. A matrix-free PDHG solver makes the resulting dense kernel-space subproblem practical on GPU hardware, while an ensemble-level exact-penalty merit function globalizes the step by jointly accounting for objective decrease, nonlinear feasibility, and particle diversity.

Across five constrained motion-planning benchmarks, SteinSQP consistently produced fully feasible ensembles while retaining the diversity benefits of particle-based inference. The experiments show that this combination of curvature, direct inequality handling, and ensemble globalization is essential: first-order constrained Stein updates can remain diverse but lose feasibility on difficult robot-scale tasks, whereas independent multistart nonlinear programming can find good single solutions but pays a serial cost and may collapse distributionally. SteinSQP therefore provides a practical middle ground between constrained optimization and constrained sampling, returning a set of feasible, low-cost, and qualitatively distinct motions in a single batched solve.

Looking ahead, we aim to extend SteinSQP toward sampling-based model predictive control under strict feasibility requirements. A key direction is to study ensemble warm starts, where the feasible particle set from the previous MPC cycle is shifted forward and refined under updated state, environment, and contact constraints, potentially reusing kernel structure and SQP dual information to support real-time constrained control.

\section*{Acknowledgments}

This work was supported by the DFG Emmy Noether Programme (CH 2676/1-1), the EU Horizon Europe projects MANiBOT (101120823) and ARISE (101135959), the BMFTR project RIG (16ME1001), and the ERC project SIREN (101163933). We also acknowledge support from the hessian.AI Service Center (BMFTR, 16IS22091), the hessian.AI Innovation Lab (S-DIW04/0013/003), TAM, RAI, Google, and the Alfried Krupp Foundation.
\appendix

\subsection{Solver Hyperparameters}
\label{app:hyperparameters}

Defaults shared across all experiments unless stated: damping
$\lambda=10^{-4}$ ($10^{-3}$ on panda); ensemble line search $K=6$,
$\beta_{\mathrm{ls}}=0.5$, $c_1=10^{-4}$, bold-driver growth $1.2$;
exact-penalty floor $\nu_{\min}=0.1$--$10$ (per example), safety
$\rho_\nu=2$; SOC damping (relative) $10^{-3}$; PDHG $L=24$--$80$ inner
iterations, $\theta=1$, safety $0.75$--$0.95$, four power-iteration sweeps
for the operator norm; basin-escape resampling every $T_r=25$--$40$
iterations with noise $\sigma_r=0.1$--$0.2$; constraint tolerance $10^{-4}$
($5\times10^{-4}$ on surface; task-scale feasibility predicates on panda and
push-box).  The diversity temperature is the fixed $\gamma=1$ in every
experiment; the online annealing schedule ($r_0=1$, $r_{\min}=0.05$,
$\beta_r=0.9$) is implemented but disabled throughout---an ablation across
the examples found it is never a net win and destabilizes the stiff contact
rollout. The code will be released upon official acceptance of this work.

\subsection{Benchmark Problem Definitions}
\label{app:nlp_problems}

Every task is an instance of the constrained least-squares NLP of
\Cref{sec:problem_definition}, solved \emph{jointly} by an ensemble of $N$
particles:
\begin{equation}
\label{eq:app_nlp}
\begin{aligned}
\min_{x\in\mathbb{R}^n}\quad
& V(x)=\frac{1}{2}\|r(x)\|^2,\\
\text{s.t.}\quad
& h(x)=0,\qquad g(x)\le 0 .
\end{aligned}
\end{equation}
Here a single decision vector $x$ encodes one candidate motion.
\Cref{tab:app_sizes} lists the problem sizes. The per-task residuals,
equalities, and inequalities are given below. For the spline tasks, the
decision vector is the flattened control polygon
$x=\operatorname{vec}(C)$, with $C\in\mathbb{R}^{M\times d}$, and the sampled
path is the deterministic image of $x$ under a fixed clamped cubic B-spline
operator. We write
\begin{equation}
\Delta^2 C_i = C_{i+2}-2C_{i+1}+C_i
\end{equation}
for the second difference of the control polygon.

\begin{table}[t]
\centering
\small
\setlength{\tabcolsep}{4pt}
\caption{Benchmark problem sizes. $N$: ensemble particles; $n$: decision-vector
dimension; $m_h$/$m_g$: number of scalar equality / inequality constraints.}
\label{tab:app_sizes}
\begin{tabular}{@{}lrrrr@{}}
\toprule
Task & $N$ & $n$ & $m_h$ & $m_g$ \\
\midrule
Annulus    & 88 & 2   & 0  & 2    \\
Trajectory & 16 & 66  & 4  & 1180 \\
Surface    & 32 & 48  & 4  & 550  \\
Panda      & 16 & 128 & 16 & 1970 \\
Push-box   & 16 & 32  & 3  & 0    \\
\bottomrule
\end{tabular}
\end{table}

\paragraph{Annulus ($n{=}2$, $N{=}88$).}
A diagnostic sampler: each particle is a point $x\in\mathbb{R}^2$, and the
objective is the negative log-density of
$\mathcal{N}(\mu,\sigma^2 I)$, with $\mu=(5,0)$ and $\sigma=2$. The feasible
set is the annulus $2.5\le \|x\|\le 3$:
\begin{equation}
\begin{aligned}
r(x) &= \frac{x-\mu}{\sigma},\\
g_1(x) &= r_{\mathrm{in}}^2-\|x\|^2,\\
g_2(x) &= \|x\|^2-r_{\mathrm{out}}^2,
\end{aligned}
\end{equation}
with $r_{\mathrm{in}}=2.5$ and $r_{\mathrm{out}}=3$. There are no equality
constraints. The mean lies outside the band, so the unconstrained optimum is
infeasible and the target is the truncated Gaussian.

\paragraph{Trajectory ($n{=}66$, $N{=}16$).}
$M=33$ planar control points define a cubic B-spline sampled at $S=120$ points
$p_s$. The objective penalizes control-polygon acceleration, the endpoints are
pinned, and each interior sample must clear $O=10$ disc obstacles:
\begin{equation}
\begin{aligned}
V(x)
&=
\frac{w_{\mathrm{sm}}}{2}
\sum_{i=0}^{M-3}\|\Delta^2 C_i\|^2,\\
h(x)
&=
\bigl(C_0-x_{\mathrm{s}},\;
      C_{M-1}-x_{\mathrm{g}}\bigr).
\end{aligned}
\end{equation}
Here $w_{\mathrm{sm}}=1$, $x_{\mathrm{s}}=(-5,0)$, and
$x_{\mathrm{g}}=(5,0)$. The collision inequalities are
\begin{equation}
g_{sj}(x)
=
(\rho_j+\delta)^2-\|p_s-o_j\|^2
\le 0,
\end{equation}
for $s=1,\ldots,S-2$ and $j=1,\ldots,O$, with margin $\delta=0.01$.
The two endpoint samples coincide with the pinned control points and are
dropped, leaving $m_g=118\times 10$ inequalities.

\paragraph{Surface ($n{=}48$, $N{=}32$).}
$M=24$ planar control points define a path lifted onto a smooth RBF terrain
$z=f(x,y)$, with $P(t)=(p(t),f(p(t)))$. The objective integrates
control-polygon acceleration and lifted 3-D path length by Gauss--Legendre
quadrature:
\begin{equation}
V(x)
=
\frac{1}{2}
\int
\left(
w_{\mathrm{sm}}\|C''(t)\|^2
+
w_{\ell}\|P'(t)\|^2
\right)
\,dt .
\end{equation}
We use $w_{\mathrm{sm}}=1$ and $w_{\ell}=4$. The endpoints are pinned by
\begin{equation}
h(x)
=
\bigl(C_0-x_{\mathrm{s}},\;
      C_{M-1}-x_{\mathrm{g}}\bigr),
\end{equation}
with $x_{\mathrm{s}}=(-4.5,0)$ and $x_{\mathrm{g}}=(4.5,0)$.
The path avoids $H=5$ spherical holes through
\begin{equation}
g_{sj}(x)
=
(r_j+\delta)^2-\|P_s-b_j\|^2
\le 0,
\end{equation}
for $s=1,\ldots,S$ and $j=1,\ldots,H$, with margin $\delta=0.05$.
This gives $m_g=110\times 5$ inequalities.

\paragraph{Panda throw ($n{=}128$, $N{=}16$).}
A $T=16$-knot joint trajectory with $7$ arm joints per knot is stored in an
$8$-wide block, giving $x\in\mathbb{R}^{128}$ with $112$ active variables.
Let $q_k\in\mathbb{R}^7$ be the knot configuration and $p_k$ the
forward-kinematic TCP position. The ball is released at the last knot with
\begin{equation}
r=p_{T-1},
\qquad
v=\frac{p_{T-1}-p_{T-2}}{\Delta t},
\end{equation}
and the flight time $t_f$ is eliminated analytically, assuming release above
the basket. The objective smooths joint acceleration:
\begin{equation}
V(x)
=
\frac{1}{2}
\sum_{k=0}^{T-3}
\|q_{k+2}-2q_{k+1}+q_k\|^2 .
\end{equation}
The equality constraints enforce the start pose, zero initial velocity, and
ballistic landing at the basket:
\begin{equation}
\begin{aligned}
h_1(x) &= q_0-q_{\mathrm{start}},\\
h_2(x) &= q_1-q_0,\\
h_3(x) &= r_{xy}+t_f v_{xy}-\mathrm{basket}_{xy}.
\end{aligned}
\end{equation}
Together these give $m_h=16$ scalar equalities. The inequalities consist of
joint limits, flight-time bounds, arm--obstacle clearance, and thrown-ball
clearance. Specifically, they include $2\times7\times16=224$ joint-limit
constraints, two flight-time bounds, $16\times13\times8=1664$ arm--obstacle
clearance constraints, and $10\times8=80$ thrown-ball clearance constraints,
for a total of $m_g=1970$. All clearance constraints use margin
$m=0.03\,\mathrm{m}$; the optimizer adds a $0.02\,\mathrm{m}$ slack.

\paragraph{Push-box ($n{=}32$, $N{=}16$).}
$M=16$ pusher control points define a smooth pusher path over $T=48$ rollout
steps. The box pose $\mathrm{box}_k\in SE(2)$ is not a decision variable; it is
obtained by the smoothed quasi-static contact rollout
\begin{equation}
\mathrm{box}_{k+1}
=
\mathrm{box}_{k}
+
h\,\mathrm{twist}(\mathrm{box}_{k},P_k),
\qquad h=0.1,
\end{equation}
with $\mathrm{box}_0=q_0=(0,0,0)$. The objective smooths the pusher control
polygon:
\begin{equation}
V(x)
=
\frac{w_{\mathrm{sm}}}{2}
\sum_{i=0}^{M-3}
\|\Delta^2 C_i\|^2,
\qquad
w_{\mathrm{sm}}=1 .
\end{equation}
The terminal box pose is pinned by
\begin{equation}
h(x)
=
\left(
\mathrm{box}_{T}^{xy}-\mathrm{goal},\;
\sin\!\bigl(2(\theta_T-\theta_{\mathrm{goal}})\bigr)
\right),
\end{equation}
where $\mathrm{goal}=(1.4,0.6)$ and $\theta_{\mathrm{goal}}=0$.
There are no inequality constraints. The heading residual is
$90^\circ$-symmetric because the rounded box is invariant under quarter turns,
which admits the distinct $\{0,\pm90^\circ\}$ rotation modes observed in
\Cref{sec:experiments_diversity}.

\bibliographystyle{IEEEtran}
\bibliography{reference}

\end{document}